\documentclass[journal]{IEEEtran}

\usepackage{cite}
\usepackage[pdftex]{graphicx}
\graphicspath{/images/} 
 
\usepackage{amsmath}
\usepackage{algorithmic}
\usepackage{array}
\usepackage{booktabs}
\usepackage{multirow}
\usepackage{multicol}
\usepackage{arydshln}
\usepackage{subcaption}
\usepackage{url}

\begin{document}
%
\title{Haris: an Advanced Autonomous Mobile Robot for Smart Parking Assistance}
%
%
%

%

\author{
  \IEEEauthorblockN{
        Layth Hamad \IEEEauthorrefmark{1} \IEEEauthorrefmark{2},
  	Muhammad Asif Khan\IEEEauthorrefmark{1},
  	Hamid Menouar\IEEEauthorrefmark{1},
        Fethi Filali\IEEEauthorrefmark{1}, and
       Amr Mohamed\IEEEauthorrefmark{2}
  }

  \IEEEauthorblockA{
  	Qatar Mobility Innovations Center \IEEEauthorrefmark{1},
   Department of Computer Science and Engineering, Qatar University\IEEEauthorrefmark{2},\\
        \{laythh, muhammada, hamidm, filali\}@qmic.com\IEEEauthorrefmark{1}, 
        amrm@qu.edu.qa\IEEEauthorrefmark{2},
  }
  }

%



\maketitle

\begin{abstract}
This paper presents Haris, an advanced autonomous mobile robot system for tracking the location of vehicles in crowded car parks using license plate recognition. The system employs simultaneous localization and mapping (SLAM) for autonomous navigation and precise mapping of the parking area, eliminating the need for GPS dependency. In addition, the system utilizes a sophisticated framework using computer vision techniques for object detection and automatic license plate recognition (ALPR) for reading and associating license plate numbers with location data. This information is subsequently synchronized with a back-end service and made accessible to users via a user-friendly mobile app, offering effortless vehicle location and alleviating congestion within the parking facility. The proposed system has the potential to improve the management of short-term large outdoor parking areas in crowded places such as sports stadiums.
The demo of the robot can be found on \url{https://youtu.be/ZkTCM35fxa0?si=QjggJuN7M1o3oifx}.
\end{abstract}



\section{Introduction}
With the increasing urbanization, our cities continue to expand with the development of new large-scale facilities such as stadiums, malls, parks, and universities. A common challenge in these large facilities is the efficient management of large car parking. Also, when organizing mega events, event-organizing agencies face severe challenges in managing the crowds \cite{khan2022revisiting, khan2023visual} and meeting the short-term parking demands. For instance, during the recent Football World Cup 2022, Qatar hosted around three million visitors and constructed eight (8) large stadiums with accompanying parking areas with a capacity of 18,000 to 20,000 cars. In such large parking facilities and particularly in outdoor parking, it is not very unlikely for many people to forget where they parked their cars, leading to inconvenience and frustration.
\par
\begin{figure*}[]
\centering
\subcaptionbox{\scriptsize The parking assistant electric scooter.}{
\includegraphics[width=4cm, height=4cm]{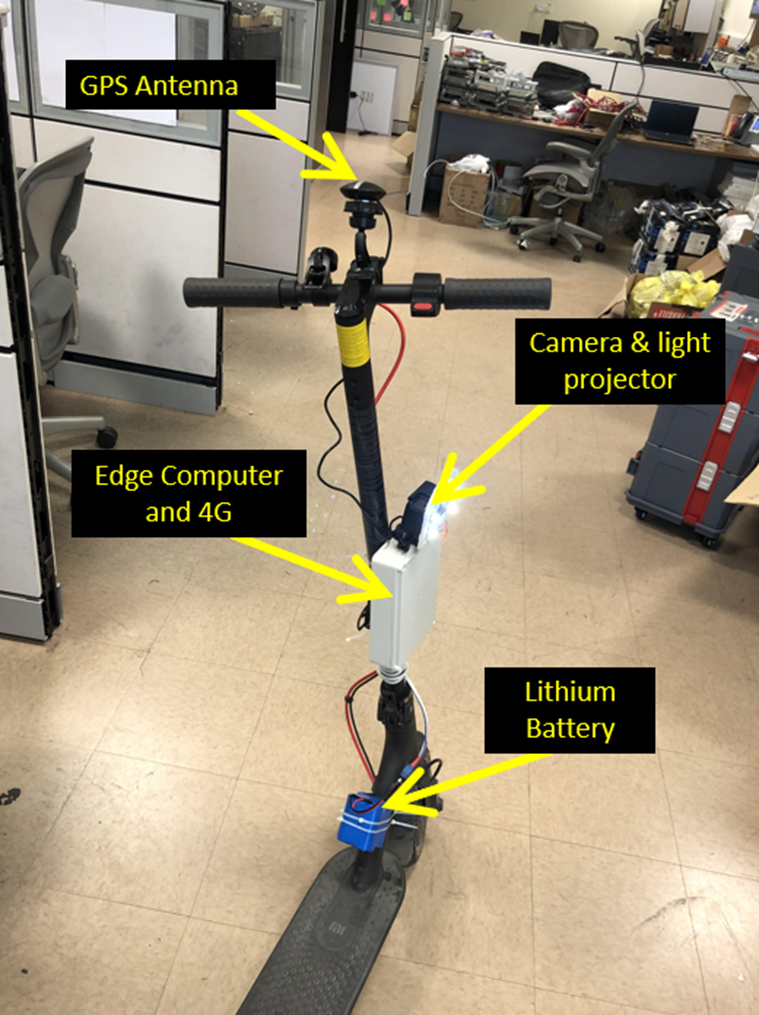}}
\hspace{3em}
\subcaptionbox{\scriptsize The fully autonomous robot (Haris).}{
\includegraphics[width=6cm, height=4cm]{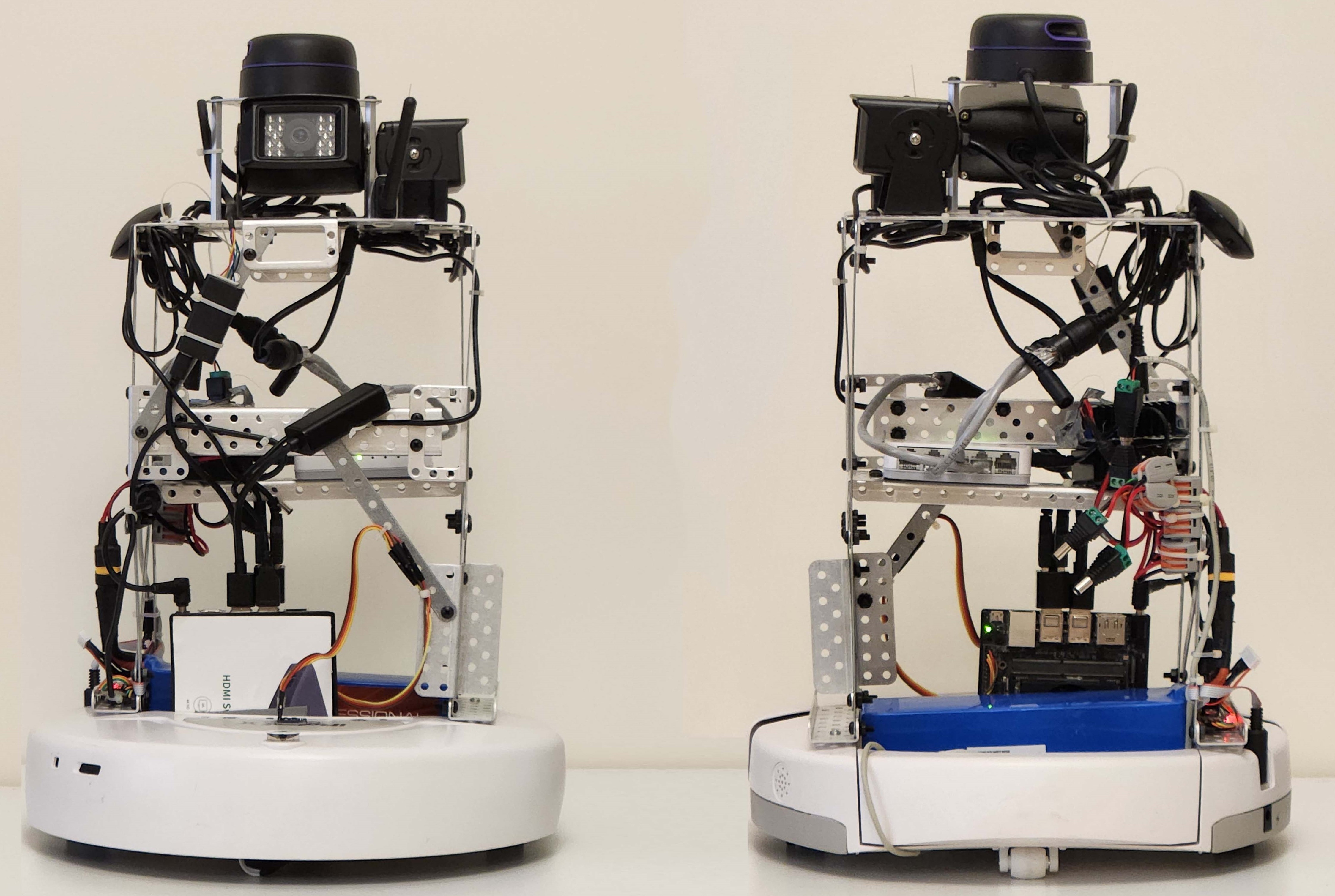}}
\caption{The structure of the proposed autonomous mobile robot (Haris) for smart parking assistance.}
\label{fig:robot_actual_pic}
\end{figure*}

To address this problem, we propose the use of intelligent mobile robots as autonomous parking assistants. The idea was the proposed autonomous robot was borrowed from our earlier solution for our parking assistant electric scooter equipped with cameras which are manually operated by an operator to scan the parking lots and record the vehicle positions. The proposed robot equipped with various sensors can autonomously navigate a large parking area without the need for an operator, scan, and recognize license plate numbers, and then store this information, along with the GPS location of the car on a backend server. This information can then be accessed by car owners through a website or a mobile application, allowing them to locate their parked vehicles easily. To achieve this goal, we leverage sophisticated computer vision algorithms, mapping and navigation techniques, and perception systems, including cameras and Light Detection and Ranging (LiDAR) to build an autonomous mobile robot. The proposed autonomous parking robot is capable of efficiently assisting with parking management and improving the visitor's experience in large parking areas. The electric scooter with a parking assistance system and the autonomous mobile robot are showcased in Fig. \ref{fig:robot_actual_pic}.

Few commercially available autonomous robots for parking assistance are recently introduced e.g., in \cite{smp_2016}. The robot is originally designed to accomplish generic autonomous tasks but is customized for ALPR using a multiple-cameras system for object avoidance. However, this system has a few limitations. First, it uses a high-power computer system for running computationally extensive tasks. Second, the robot doesn't implement any path-planning algorithm. Our proposed autonomous robot system offers enhanced features using an integrated AI system such as object detection and avoidance, pre-planned mission, and path-planning algorithm. There are other similar products commercially available in the market however, all of these use GPS-based navigation. These robots might not be suitable for many scenarios with high accuracy requirements due to the inherited limitation of GPS i.e., +-10 meters average error. Our proposed mobile robot (a.k.a. Haris) overcomes these limitations with a more accurate yet reliable navigation system with less than 5cm location error.
\par
The robot streams self-generated GPS-like coordinates (latitude and longitude) to mimic the actual location on earth, based on the movement of the robot in the robot world coordinate system (relative distance from the initial point). We developed an algorithm that uses a bidirectional transformation method to continuously translate robot trajectory to GPS coordinate points (lat, long), and allow the robot to execute missions sent as a set of waypoints destinations represented in GPS coordinates. The transformation process is initialized and synced based on a fixed reference point (usually the robot's stationary charging stations), where the algorithm tries to reduce the accumulated error by re-initialize the transformation parameters each time the robot reaches a reference point (charging station).

The contribution of our work is as follows:
\begin{itemize}
\item We developed a highly sophisticated and reliable autonomous robot for parking assistance without relying on a GPS system for navigation. The robot scans and maps the locations of vehicles in large-scale parking areas.
\item The robot uses a reliable framework to integrate computer vision models for robust automatic license plate recognition (ALPR) using optical character recognition (OCR) for reading plate numbers.
\item The proposed system has been tested successfully in one of the corridors of a Qatar FIFA World Cup stadium serving around 10,000 parking spaces.
\end{itemize}


\section{Related Work} \label{sec:related-work} 
Autonomous driving robots have gained a lot of attention recently due to various factors, especially since the fast-paced advances in artificial intelligence and computer vision. In \cite{812788}, authors developed an autonomous mobile robot designed to track humans using face detection. Multiple detection approaches are used for face detection e.g., color, and contour information. Once a person is detected, a stereo algorithm is used to specify the person's position. 

In \cite{9712151}, authors developed a mobile robot platform with multi-target recognition and tracking. The system is designed to accurately detect and track multiple moving targets in real-time for smart surveillance in smart cities. The authors proposed a mobile edge computing (MEC) based mobile robot system to control the motion of the robot to run to various positions according to the target detection. The system uses positioning data and identification of path signs. The robot is designed to perform surveillance by classifying the detected object and triggering an alarm when an unusual activity is detected. The data is stored and processed at the MEC server. The MEC-based data storage and computing help alleviate the processing latency. However, the detection and tracking accuracy of moving targets is limited and the system performance degrades at various complex tasks such as activity recognition and behavior analysis.In \cite{Connie2018} authors developed a mobile robot for an automatic license plate recognition system (ALPR). However, the accuracy of the system was less than a similar system with fixed cameras.
\par
A closely related work \cite{6154014} proposes an Android-based robot featuring automatic LPR and automatic license plate patrolling (LPP) systems. The LPR identification system incorporates four techniques: Wiener deconvolution vertical edge enhancement, AdaBoost plus vertical edge license plate detection, vertical edge projection histogram segmentation (VEPHS) stain removal, and customized optical character recognition. For LPP, three further techniques are employed i.e., HL2-band rough license plate detection, orientated license plate approaching, and Ad-Hoc-based remote motion control. The automatic robot starts with an auto patrol, then detects the license plate if the plate number exists, performs edge enhancement, does license plate localization, and then extracts the character and matches the license plate in a database.

\section{Proposed System Design} \label{sec:system-design} 
\begin{figure*}[]
\centering
\includegraphics[width=0.55\textwidth]{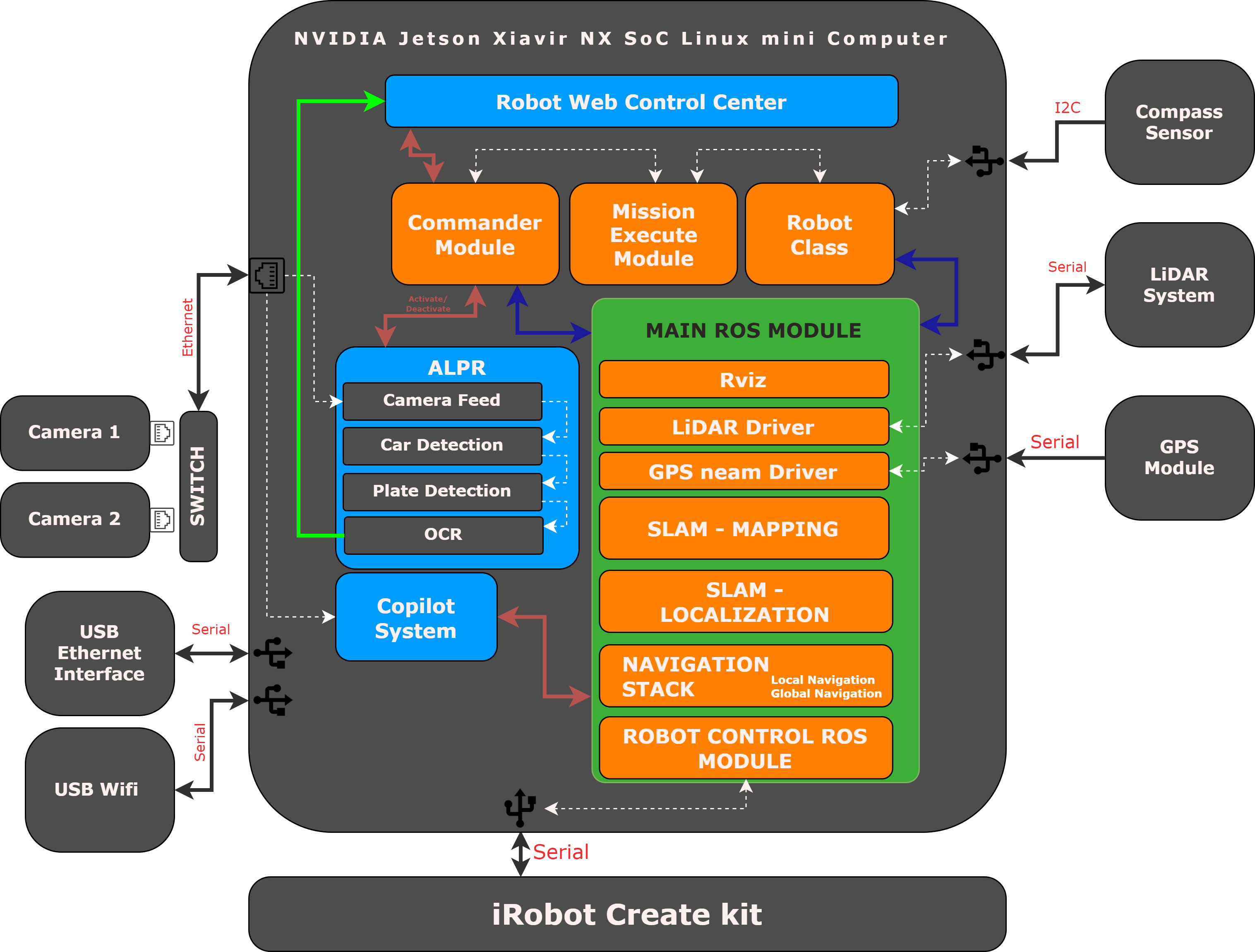} \caption{Architecture of the proposed system.}
\label{fig:robot_block_all}
\end{figure*}

A high-level architecture of the proposed system is depicted in Fig. \ref{fig:robot_block_all}. The detailed design and features of each module are described in the following sub-sections.

\subsection{Mobile Robot Hardware System}
\begin{figure}[htbp]
\centering
\includegraphics[width=0.9\columnwidth]{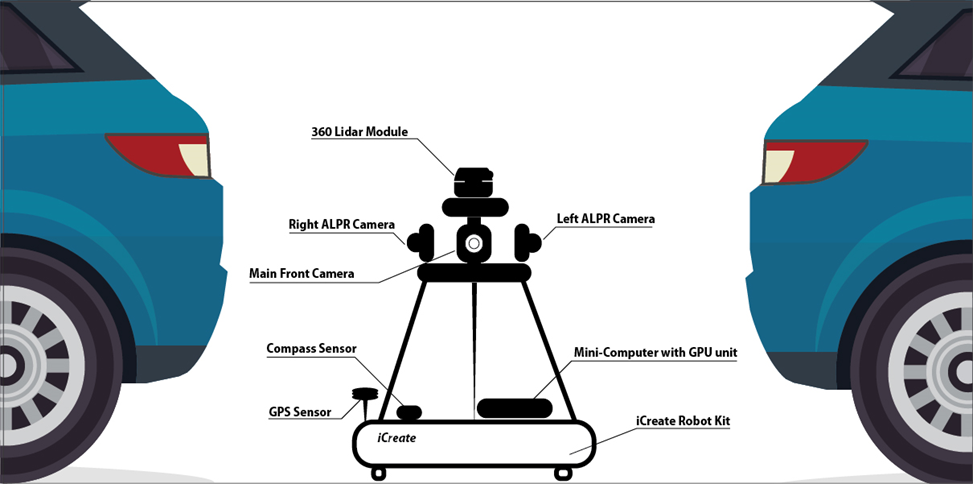} 
\caption{Mobile Robot System Architecture}
\label{fig:robot_archit}
\end{figure}

The mobile robot is a three-wheeled structure with two driven wheels. To create a solid foundation for our robot, we used the iRobot Create robot kit (a.k.a. the Romba robot) as the base. This kit provided the necessary mobility and sensing hardware components in an efficient manner, allowing us to focus on designing and implementing other parts of the robot. The general layout of our robot is shown in \ref{fig:robot_archit}.

\subsubsection{iRobot Create Robotic kit}
We utilized an off-the-shelf robot kit \textit{iRobot Create kit} that offers the necessary hardware capabilities for mobility (including motors and a driver board) and various types of sensors. This kit allows for easy communication with these components, enabling us to concentrate on the desired functionality and task at hand rather than spending time and resources on designing the motor circuits and electronics.

\subsubsection{The Power system}
The main power system is a 12V battery which provides the power to all subsystems including iRobot internal components to drive the motors and other sensors/actuators (19.2V), the mini-computer unit (19.2V), the camera system, 2 cameras are used in this approach (12V), the network switch (5v), and the external GPS and compass modules(5v and 3.3v from The mini-computer unit).

\subsubsection{The Mini-Computer Unit}
The proposed robot is designed to be an autonomous driving robot. To facilitate this capability, a high-performance computing unit is necessary to run the AI and computer vision algorithms. We have selected the NVIDIA Xavier NX Development Kit as the computer unit for our robot as in \cite{jablonski2021evaluation}.

\subsubsection{The GPS Module}
GPS is useful but not an essential module of our system. It is used to set the initial location of the robot, after which the robot is able to self-navigate. However, GPS has a margin of error of approximately 10 meters, which is a problem for our robot as it is only 60cm in size and cannot operate accurately with such a large error margin. To address this issue, we have selected the \textit{Globalsat BU-353S4} USB GPS Receiver to get reliable GPS data and easily interface with the main computer unit through USB.
\subsubsection{The Compass Module}
The iRobot Create does not have a compass sensor, so we have installed the \textit{HMC5883L GY271} compass sensor to determine the accurate direction of the robot in relation to the global north of the earth. This initial angle is used in the transformation between the robot's local coordinate system and the global GPS system (represented in latitude and longitude). 
\subsubsection{The Cameras and the Local Network System: }
Two high-quality, high-fps cameras connected to a local network switch are used that support the Real Time Streaming Protocol (RTSP) and can be accessed by various modules. The front camera will be used for the object detection copilot system, while the side-facing camera will be used for the license plate recognition system.

\subsubsection{The LiDAR Module}
LiDAR (Light Detection and Ranging) can be used in SLAM \cite{bailey2006simultaneous} to create a map of the environment and simultaneously determine the location of the robot within that map. By measuring the distance between the robot and the object for a number of points in the environment, a 2D map of the surrounding area is generated.

\subsubsection {ALPR module}
The ALPR module detects and reads plate numbers, this module consumes the side camera stream and processes the frames to detect and read all plate numbers.

\subsubsection {The Communication module}
For this particular module, tasked with facilitating communication between ROS and non-ROS modules, we have chosen Message Queuing Telemetry Transport (MQTT) as the preferred communication protocol. This decision is based on MQTT's proven effectiveness and reliability in enabling seamless data exchange between disparate system components

\subsection{Mobile Robot Software System}
The software system consists of the following components.

\subsubsection{The Robot Operating System (ROS)}
The Robot Operating System (ROS) serves as the main backbone of the system, running the required modules to perform several tasks, mainly the SLAM and navigation. The ROS is a flexible framework for writing robot software. It is a collection of tools, libraries, and conventions that aim to simplify the task of creating complex and powerful robotics systems. ROS is structured as a distributed system, meaning that its functionality is divided into smaller units called nodes, which communicate with each other using a publish-subscribe messaging paradigm. Nodes can be written in any programming language, as long as they can communicate with the rest of the system using the ROS standard message formats and network protocols \cite{basheer2019overview}.

The ROS architecture consists of several main components including the ROS Master, nodes (for sensing, actuation, or data processing), topics (one-way communication channels), and services (two-way communication channels).

In addition to these core components, ROS also provides a collection of libraries and tools for common robotics tasks, such as robot simulation, robot navigation, and robot manipulation. 

\begin{itemize}
\item \textit{Robot ROS Transformation Tree: } In ROS, a transformation tree is a data structure that is used to represent the relationships between different coordinate frames in a robot system. A coordinate frame is a reference frame that is used to specify the position and orientation of an object in space. The transformation tree defines the relationships between these coordinate frames, allowing nodes in the system to convert coordinates from one frame to another \cite{ogiwara2022making}. The transformation tree is typically represented as a directed acyclic graph (DAG), with the coordinate frames as the nodes and the transformations between the frames as the edges. The root node of the DAG represents the global coordinate frame, which is typically fixed with respect to the robot's environment. The other nodes in the DAG represent the local coordinate frames of the various sensors, actuators, and other components of the robot.

\item \textit{ROS USB LiDAR module: } It is a ROS module that provides access to LiDAR (Light Detection and Ranging) sensor data through a USB interface. LiDAR sensors are used to measure distance and create 2D or 3D maps of the environment by emitting laser beams and measuring the time it takes for the beams to reflect off objects. We used the LiDAR system as the main source of surrounding perception, where other nodes in the system are designed to process and interpret the LiDAR data, such as for obstacle detection or mapping.

\item \textit{ROS GPS NEAM driver: }
The GPS driver consumes the data from the GPS module through USB, the data is published in the form of GPS coordinates (latitude and longitude), altitude, and timing information. A GPS NEAM (North, East, Altitude, and Map) driver is a specific type of GPS node that is designed to provide data in a local NEAM coordinate system, rather than the global GPS coordinate system.

\item \textit{ROS odometry module: }
The odometry module is a software component that estimates the position and orientation of a robot based on its movement over time. Odometry is a common technique for localizing robots, particularly in situations where other forms of localization, such as GPS or LiDAR, are not available or are too noisy. Our odometry data is being produced by the iRobot Create kit, and being red by the iRobot Create driver.

\item \textit{ROS iRobot Create Driver: }
The ROS iRobot Create driver is a software module that provides a ROS interface for the iRobot Create. The driver allows users to control the robot and access its sensor data using ROS topics and services. Typically, the ROS iRobot Create driver consists of several nodes that handle specific tasks, such as low-level communication with the robot's microcontroller, sensor data processing, and high-level control. The driver may also provide additional features, such as support for mapping and localization, or integration with other ROS packages.
\end{itemize}

\subsubsection {The computer vision AI algorithm}
A highly sophisticated and lightweight CNN model detects all objects in front of the robot, and the module consumes the stream from the main front camera and processes each frame to detect the object in that frame. 

\subsubsection {Robot Control Web App}
This module has been built for users to be able to track and send missions to the robot. It includes a map and control command and works as follows. It first initializes the robot's location if the GPS data is inaccurate. Then it transmits the initial location to the robot. Finally, it creates a mission with waypoints on the map and transmits the mission to the robot, and starts it.
\subsubsection{Robot Subsystems and Drivers:}
In this section, we will outline the design of the subsystems that will enable the global task described in this report to be completed.

\subsubsection{Robot Class Module}
The Robot class is a software abstraction that represents a robot in a simulated or real-world environment. It is implemented in Python and serves as a blueprint for defining the properties and behaviors of the robot. This includes characteristics such as the robot's location, state, and movement capabilities, as well as its sensors and actuators. By defining these properties and behaviors through the Robot class, it becomes possible to design customized functionality for the robot to interact with the core ROS modules. These methods can be called by other software components to control the behavior of the robot and achieve a desired goal. Overall, the Robot class plays a central role in the robot system operation.

\subsubsection{Robot Mission Executor Module}
A Robot Mission Executor is a Python module that is responsible for executing a predefined mission or set of tasks for a robot. The mission might be a high-level goal, such as navigating to a particular location or performing a series of actions in a particular order, possibly following a set of GPS waypoints. The module is responsible for coordinating with the Robot class to execute the required goals. It has implemented two main functionalities: First, it converts the GPS waypoints (represented in latitude and longitude) to the robot's work coordinates (represented in xyz in meters), using the GPS-Robot world transformation module. Second, it sends each waypoint as the next mission to the Robot class and waits until the mission is completed before sending the next one.

\subsubsection{Robot Mission Executer Module}
The Robot Mission Executor module is responsible for coordinating with the front-end Robot Control Center Web App and other computer vision modules to execute a predefined mission or set of tasks. This coordination involves activating and deactivating various modules as needed to achieve the desired goals. For example, the module might activate the Automatic License Plate Recognition (ALPR) camera module when the robot is on a mission and moving, to gather data about the environment. When the robot is not on a mission or is stationary, the module could deactivate the ALPR camera and other modules to conserve resources and reduce computational load. The module plays a critical role in enabling the robot to carry out its mission efficiently and reliably. It acts as a central hub for coordinating the various components of the system, ensuring that the right modules are activated at the right time to achieve the desired goals.

\subsection{Algorithms}

We used Simultaneous Localization and Mapping (SLAM) \cite{bailey2006simultaneous} to create a map of an unknown environment while simultaneously determining the robot's location within that environment. SLAM is a crucial capability for many mobile robot applications, such as autonomous navigation, exploration, and localization. In ROS, SLAM is typically implemented as a collection of nodes that work together to solve the SLAM problem. These nodes may include sensors such as LiDAR or cameras, as well as algorithms for mapping, localization, and other tasks.
Furthermore, the Robot Navigation stack is built to enable mobile robots to navigate and explore their environment autonomously. The robot also uses its internal coordinate system to represent its global position (taken from GPS sensor initially) and movement. For automatic license plate recognition of vehicles, we used YOLOv7 \cite{Wang2022YOLOv7TB} for detecting cars and license plates, and OCR (Optical Character Recognition) to read the vehicle number.

\section {Experiments and Results} \label{sec:experiments}
The focus of this section is to test the robot's operations in real-world scenarios in order to evaluate its performance in handling different tasks.

\subsection{ROS LiDAR Mapping}
\begin{figure}[!h]
\centering
\includegraphics[width=0.7\columnwidth]{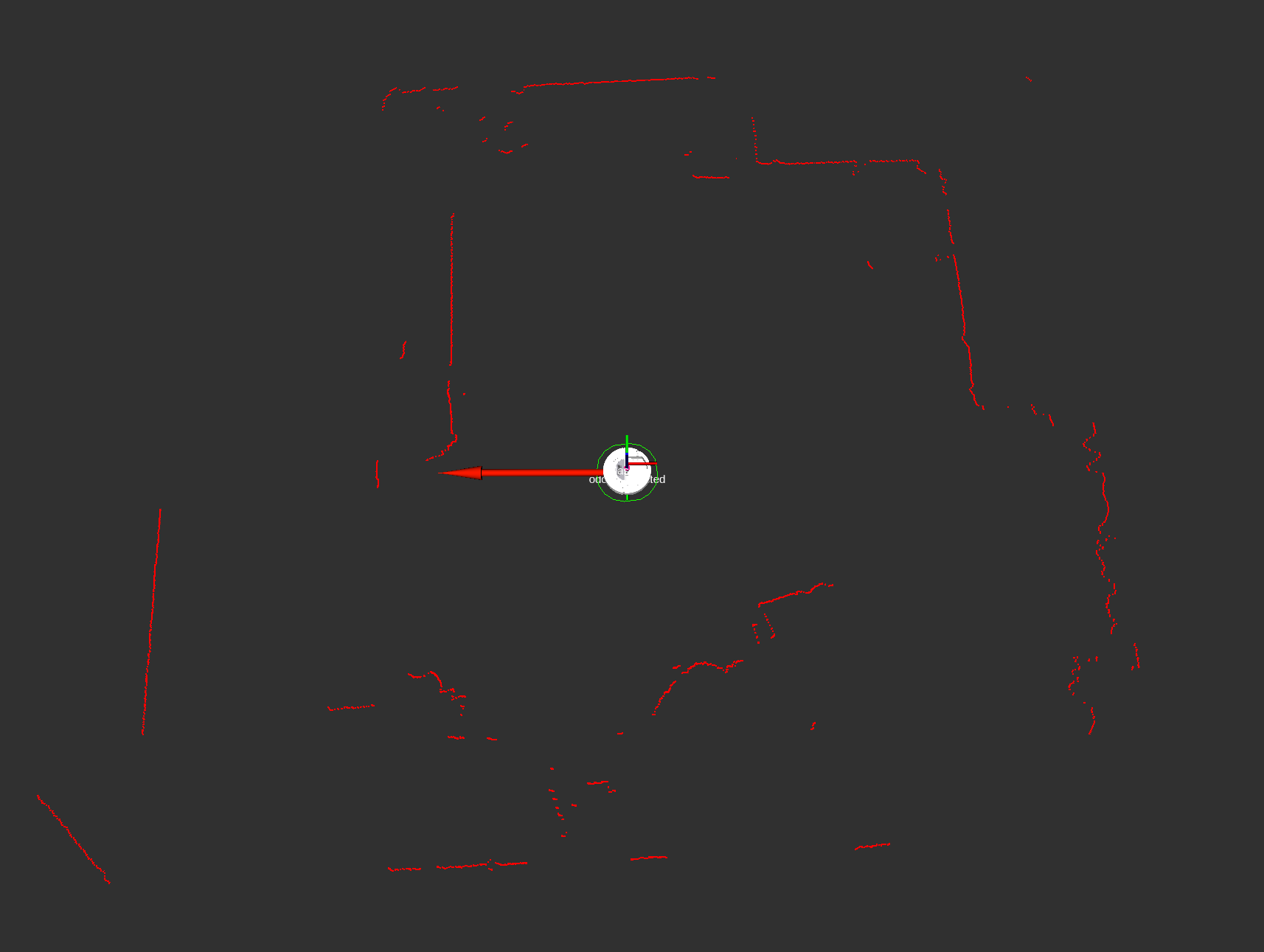} 
\caption{LiDAR system test (indoor).}
\label{fig:LiDAR_test}
\end{figure}

Fig. \ref{fig:LiDAR_test} shows an actual test of the ROS USB LiDAR node consuming the sensor data from the SLAMTEC RPLiDAR model A3. The initial results of testing the LiDAR module in an indoor environment show that the LiDAR was able to detect the surrounding objects and the shape of the room with high-quality cloud points while detecting walls (straight lines with no shifting). It also shows the ability of the robot to localize itself inside the room in the correct direction.

\subsection{Robot Mapping and Localization}
We tested two SLAM methods i.e., Gmapping and Hector SLAM. Gmapping is a ROS package that provides an implementation of the SLAM algorithm. It is based on a probabilistic approach, using a particle filter to track the robot's pose and a FastSLAM algorithm to build the map. The accuracy and performance of Gmapping rely on the accuracy of both the odometry data and the LiDAR readings. While we found the LiDAR data to be highly accurate, the odometry data was found to be insufficiently accurate. This is most likely because the low-quality sensors used in the iRobot Create Kit were unable to detect certain types of movements, such as rotations.

\begin{figure}[!h]
\centering
\includegraphics[width=5cm, height=5cm]{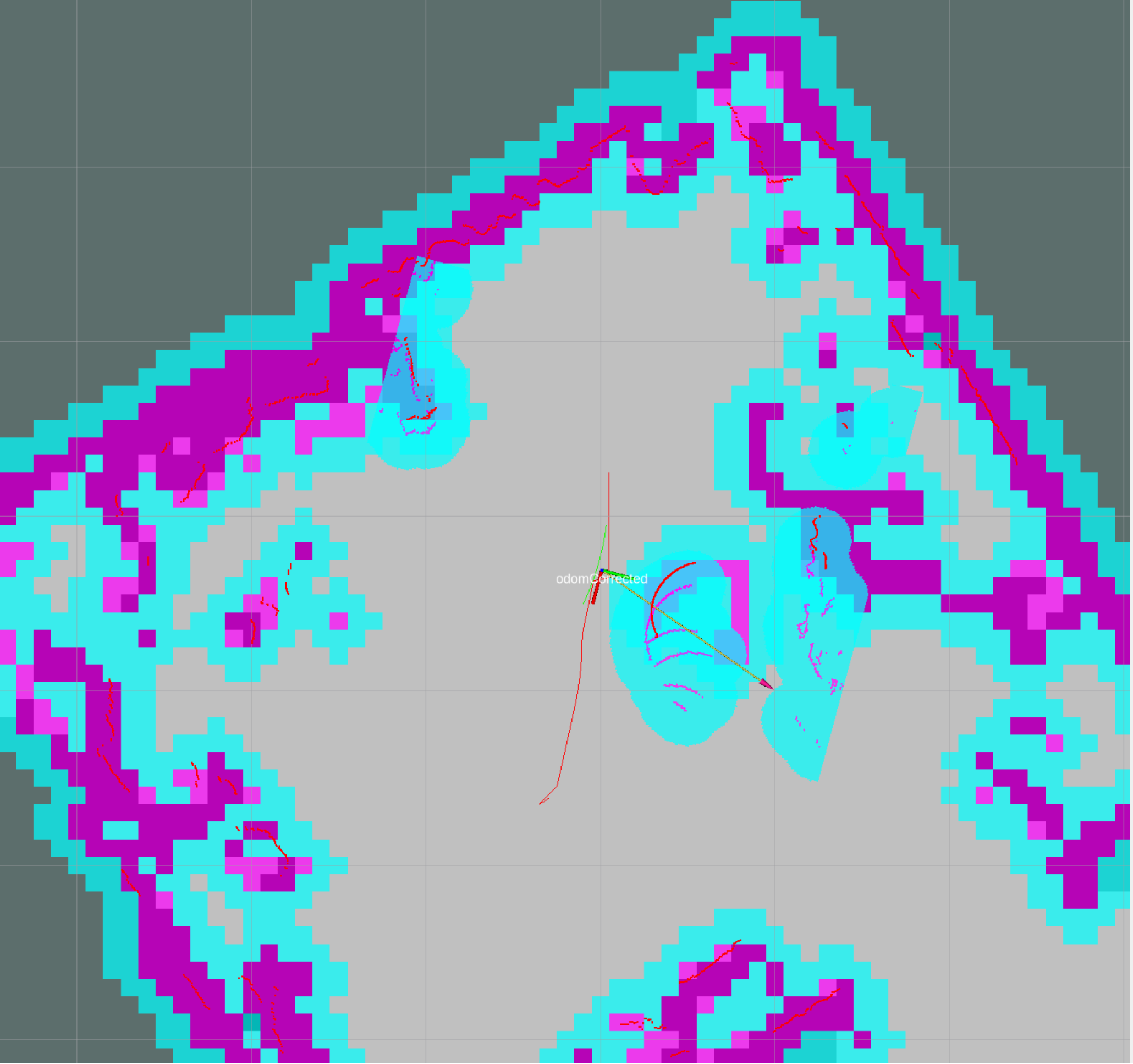} 
\caption{Robot Global and Local Navigation Plan}
\label{fig:gloab_loal_nav}
\end{figure}

Fig. \ref{fig:gloab_loal_nav} demonstrates the robot's global navigation plan (red lines), and the local navigation plan (green lines). One can observe the global trajectory (red color) representing the shortest path the robot can follow from origin to destination while avoiding the detected objects (pink).

Next, we evaluate the navigation accuracy of the proposed robot. For comparison, we conducted two sets of experiments. 

\begin{figure}[!h]
    \centering
    \subcaptionbox{\scriptsize Experiment 1: Location errors using GPS-only scheme for indoor navigation.}{
    \includegraphics[width=0.9\columnwidth]{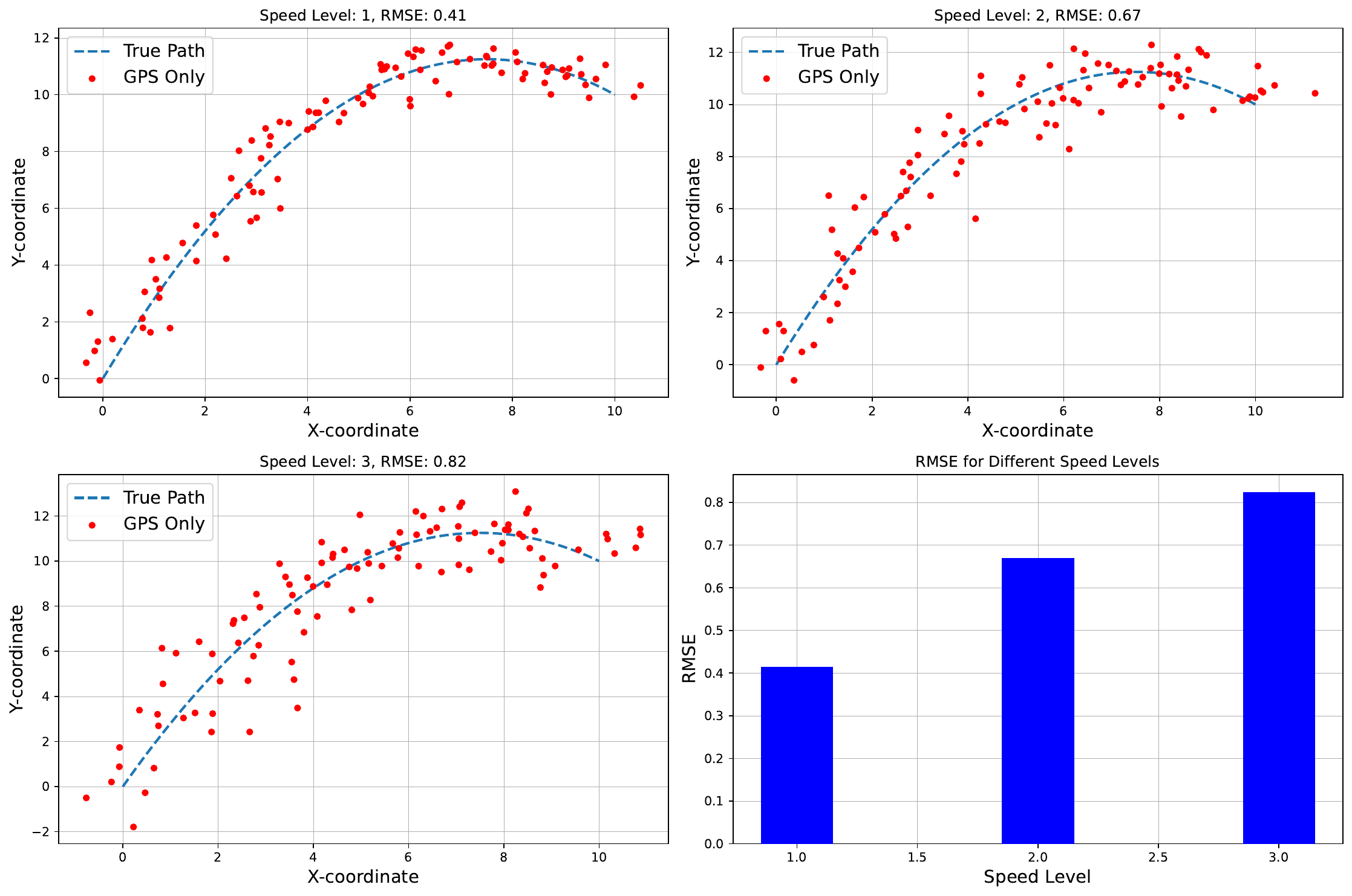}}
    \hspace{3em}
    \subcaptionbox{\scriptsize Experiment 2: Location errors using the proposed scheme for indoor navigation.}{
    \includegraphics[width=0.9\columnwidth]{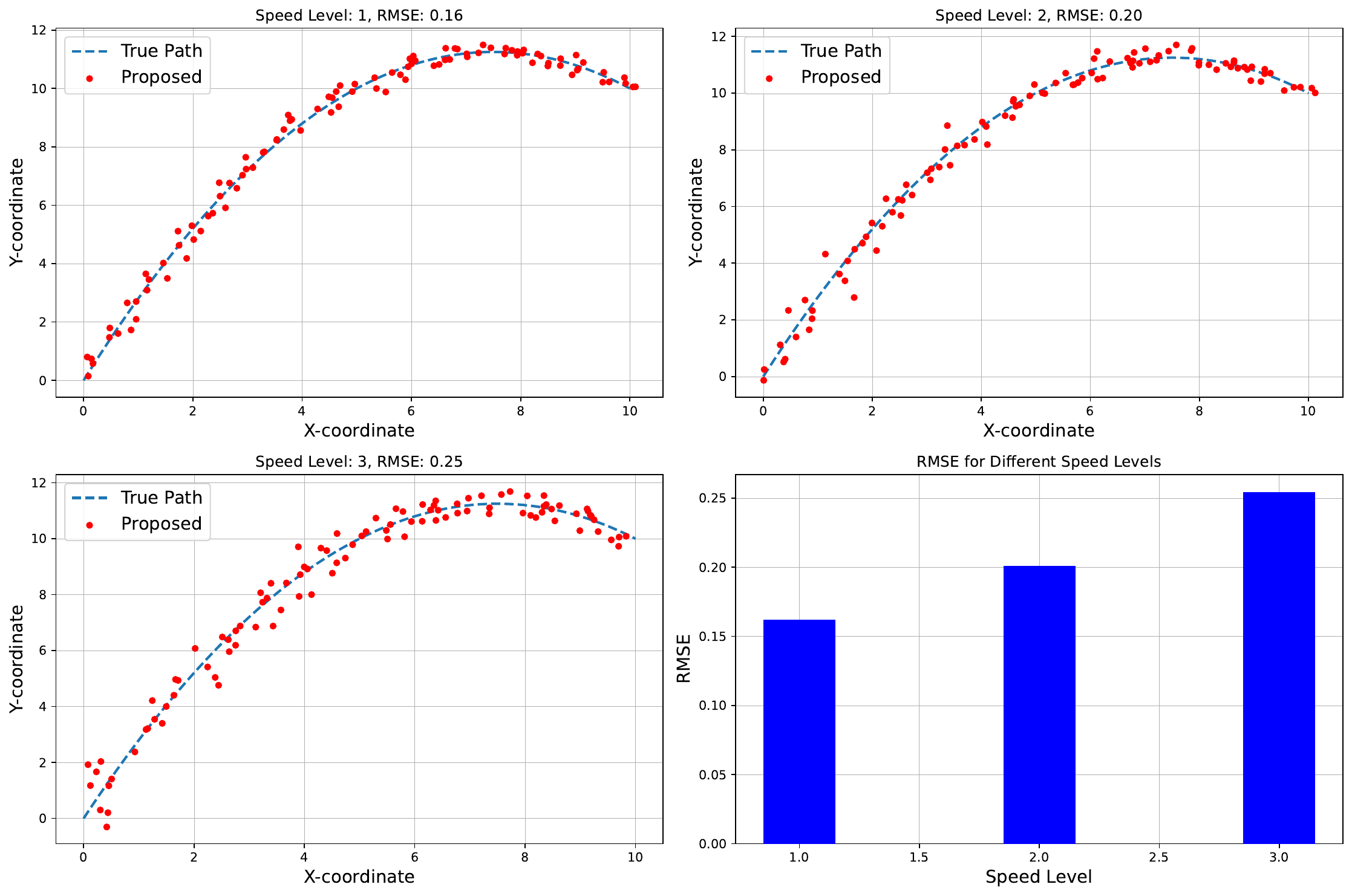}}
    \caption{A comparison of robot's path errors (RMS values) at different speed levels.}
    \label{fig:path_following}
\end{figure}

In the first experiment, we operated the robot using only a GPS system in an indoor scenario to traverse a given path. Then in the second experiment, the robot is operated with the proposed method to traverse the same path. The performance comparison is illustrated in Fig. \ref{fig:path_following}. In Fig. \ref{fig:path_following}-a (GPS-only), the error values between the actual path (straight line) and the robot path (red dots) are much higher than in Fig. \ref{fig:path_following}-b (proposed scheme), showing the accuracy of the proposed scheme.

\subsection{Detection and ALPR Recognition}
\begin{figure}[!h]
    \centering
    \includegraphics[width=0.7\columnwidth]{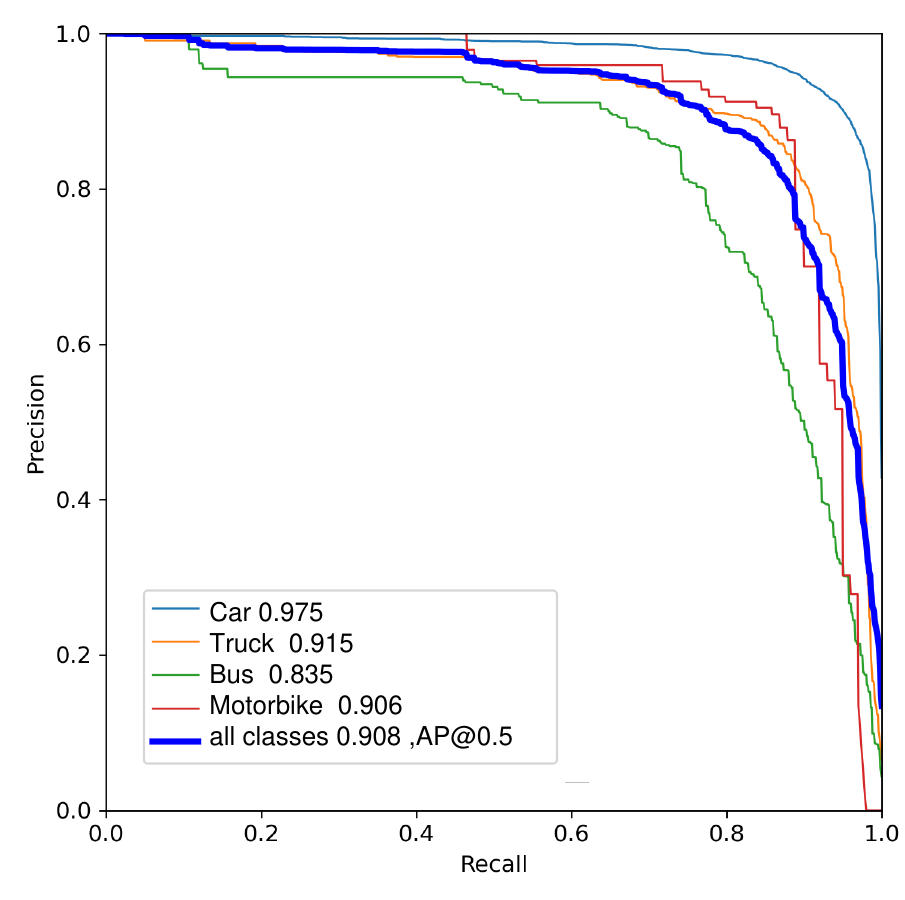}
    \caption{PR curve of the object detection model.}
    \label{fig:pr_curve}
\end{figure}

The plate number recognition system using an autonomous robot equipped with a SLAM algorithm showed promising results in its ability to scan parking lots and accurately detect and recognize plate numbers. While the system performed well during daylight hours, there was a decrease in accuracy in low-light conditions and at night, likely due to the decreased image quality captured by the camera.
Table \ref{table:detection} shows the detection performance for various vehicle types, whereas Fig. \ref{fig:pr_curve} illustrates the precision-recall (PR) curve.

\begin{table}[!h]
\centering
\caption{Car detection results.}
\label{table:detection}
\renewcommand{\arraystretch}{1.2}
\begin{tabular}{|l|c|c|c|c|c|c|}
\hline
Class & Images & Labels & \( P \) & \( R \) & \( F1 \) & \( \text{mAP@.5} \)  \\
\hline
All & 1370 & 6821 & 0.865 & 0.864 & 0.866 & 0.908 \\
Car & 1370 & 5496 & 0.905 & 0.947 & 0.926 & 0.975 \\
Truck & 1370 & 824 & 0.858 & 0.873 & 0.865 & 0.915 \\
Bus & 1370 & 402 & 0.801 & 0.771 & 0.785 & 0.835 \\
Motorbike & 1370 & 99 & 0.896 & 0.867 & 0.879 & 0.906 \\
\hline
\end{tabular}
\end{table}

The analysis of these results shows the robot's detection capabilities and highly accurate results (mean average precision i.e., mAP of 0.908) achieved using object detection and the optical character recognition module with the highest accuracy recorded for cars, followed by trucks, motorbikes, and buses respectively. 

\section*{Acknowledgement}
This work was jointly supported by Qatar University and the University of Guelph - IRCC-2023-171. The findings achieved herein are solely the responsibility of the authors.

\section{Conclusion and Future Work} \label{sec:conclusions}
This paper presented a prototype of an autonomous parking assistance robot designed to efficiently handle large parking lots and improve the user experience. The integrated robotic system utilizes SLAM techniques to navigate more reliably and accurately without relying on a GPS system. A detailed description of the design, including all hardware components, software modules, and tools, is provided, along with indoor and outdoor real-world testing to determine the prototype's effectiveness. The findings of the multiple experiments in a real environment suggest that the robot can move precisely and perform predetermined tasks, such as recognizing license plates in parking lots. Future work will involve refining the system by employing a more rugged, outdoor-suited robotic kit and evaluating its effectiveness over longer and more extensive trips. In addition, the introduction of a flying drone is an intriguing addition to the suggested system, since it provides more mobility and operational flexibility.

\bibliographystyle{IEEEtran}
\bibliography{biblio}

\end{document}